\documentclass[letterpaper]{article}

\usepackage{natbib,alifeconf}  
\usepackage{url,hyperref}
\usepackage{booktabs}
\usepackage{lipsum}
\usepackage{amsmath}
\usepackage{amssymb}
\usepackage{cleveref}

\usepackage{graphicx}
\newcommand{\emojilizard}{\includegraphics[height=1em]{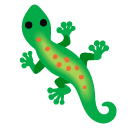}}
\newcommand{\emojismiley}{\includegraphics[height=1em]{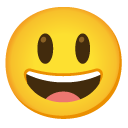}}
\newcommand{\emojispiderweb}{\includegraphics[height=1em]{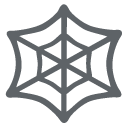}}

\defcitealias{Masumori_undated-iu}{Masumori et al., under review}

\newcommand\blfootnote[1]{%
  \begingroup
  \renewcommand\thefootnote{}\footnote{#1}%
  \addtocounter{footnote}{-1}%
  \endgroup
}

\title{Transient State Reorganization and Cell Differentiation in the Developmental Dynamics of Growing Neural Cellular Automata}

\author{
    Hiroki Sato$^{1,2,3}$,
    Atsushi Masumori$^{1,2,3}$, \and
    Takashi Ikegami$^{1,2,3}$ \\
    \mbox{}\\
    $^1$The University of Tokyo, Japan \\
    $^2$Alternative Machine Inc., Japan\\
    $^3$Atomi University, Japan\\
    hsato.ac@gmail.com
}

\begin{document}
 
\maketitle
 
\begin{abstract}
Growing Neural Cellular Automata (GNCA) develop complex morphologies from a single seed cell through shared local rules, yet the internal dynamics of this process remain poorly understood. To investigate how GNCA grows, the full developmental trajectory of trained GNCA models was traced. The trajectory of cell state development revealed that morphological convergence often proceeds non-monotonically through transient intermediate configurations. In addition, channel-wise analysis showed that the hidden channels self-organize into modular groups in parallel with the visible form. Furthermore, geometric analysis of the cell state space indicated that cell states diversify within a low-dimensional, smooth manifold. To examine cell development in more detail, community detection on an $\epsilon$-neighbour network of cells was conducted. This analysis successfully extracted discrete cell types from this continuous space, and identified transient cell-type communities during early development and stable, finer-grained types corresponding to spatially coherent regions of the mature morphology. The temporal coordination of these phenomena across multiple independent measures indicates that the developmental process of GNCA is a reorganization of transient states rather than incremental refinement.
\end{abstract}

 
\blfootnote{\textcopyright  2026 Hiroki Sato, Atsushi Masumori, Takashi Ikegami. Published under a Creative Commons Attribution 4.0 International (CC BY 4.0) license.}

\section{Introduction}
All biological organisms develop over time.
No organism is born in its completed form, but continuously constructs and reconstructs itself throughout its lifetime.
This is particularly evident in multicellular development and action, where a single fertilized egg gives rise to a diversity of cell types through local cell-cell interactions.

Growing Neural Cellular Automata (GNCA) are computational systems inspired by this paradigm.
In GNCA, its update rule is defined as a shared neural network which receives input of neighbouring cell states and outputs updated cell state. 
By training the neural network, GNCA can proliferate into global spatiotemporal patterns as a distributed asynchronous process. 
Since the introduction of GNCA by \citet{Mordvintsev2020-pp}, the framework has been extended to self-repairing morphologies \citep{Mordvintsev2020-pp}, texture synthesis \citep{Mordvintsev2021-eb}, 3D morphogenesis \citep{Sudhakaran2021-db}, soft robot co-design \citep{Horibe2021-qk,Pontes-Filho2022-et,Sato2024-jm}, and evolutionary search \citep{Palm2022-kk}.

However, analyses of the dynamics of GNCA are rare. 
Although GNCA has occasionally been compared with the development of living systems, the validity of this comparison cannot be assessed without knowing how its mechanisms work. 

Some prior work has visualized hidden channels or conducted ablation studies \citep{Greydanus2022-ic,Catrina2024-zx}. 
Also, a recent information-theoretic analysis has characterized the post-developmental dynamics of trained models \citepalias{Masumori_undated-iu}. 
However, a comprehensive developmental analysis, one that traces the full trajectory from seed to mature form, has not been undertaken. 

This study asks how a GNCA forms a complex morphology from a single seed through analyses focusing on the differentiation of internal cell states in a channel-wise and cell-wise manner during its development. 
As a highlight of the present analysis, it is shown that GNCA develops through reforming intermediate developmental structures rather than monotonically reconstructing a target state. 
Beyond the wide range of GNCA applications, the results of this study potentially contribute to establishing GNCA as a framework for studying developmental dynamics as a general phenomenon.
 
\section{Materials and Methods}
GNCA analyzed here followed the architecture introduced in the original work by \citet{Mordvintsev2020-pp}. 
In the architecture, each cell carries a 16-dimensional state vector, of which the first four channels correspond to RGBA values and the remaining twelve serve as hidden channels. 
At each time step, every cell perceives its local neighbourhood via fixed filters (identity and Sobel filters) and updates its state through a small neural network shared across all cells. 
A stochastic cell update mask is applied at each step, where each cell is updated with a probability determined by the update rate (UR).

Each GNCA model has a target image (cell state configuration). 
The shared neural network model is trained to repeatedly update cell states from a given cell states and reconstruct the target. 
In this version of GNCA, the growth begins from a single seed cell placed at the centre of the grid. 
At initialization, the alpha and hidden channels of this seed cell are set to one, while all other channels and all other cells are set to zero (i.e. black cell with channel values of one). 
A cell is treated as alive when the maximum alpha channel value within its $3 \times 3$ neighbourhood exceeds 0.1. 
Because a cell becomes alive once a neighbour reaches a sufficient alpha value, the alive region expands outward from the seed cell over successive steps. 

The analyzed GNCA models were implemented using PyTorch and trained using the Adam optimizer with a learning rate of $2 \times 10^{-3}$ and exponential decay ($\gamma = 0.9999$). 
Training proceeded for $8,000$ epochs with a batch size of $8$ drawn from a pool of $1,024$ states, with the highest-loss sample in each batch replaced by the seed state to maintain pool diversity. 
At each training iteration, the model was given a random number of time steps uniformly sampled between $64$ and $96$ to develop. 
The loss was the mean squared error (MSE) between the target RGBA image and the model output. 
In the present study, the grid was set to have circular boundary conditions. 
 
We trained models under two update rate conditions: UR = 0.5 (asynchronous updating, where each cell has a 50 \% probability of being updated at each step) and UR = 1.0 (synchronous updating, where all cells are updated at every step). 
Three target emoji images were used — \emojilizard (U+1F98E, \texttt{lizard}), \emojismiley (U+1F603, \texttt{smiley}), and \emojispiderweb (U+1F578, \texttt{spider-web}) — sourced from the Google Noto Emoji dataset at a resolution of $40 \times 40$ pixels with 16 pixels of padding, yielding a $72 \times 72$ grid (see \cref{tab:emoji_mapping} for reference).
For each combination of update rate and target emoji, eight independent runs with different random seeds were conducted, yielding $n=8$ seeds per update rate condition.
During training, circular damage masks were applied to the last three samples of each batch ($N_{\text{damage}}=3$) to encourage the self-repair capability of the trained models. 
This produced a total of 48 trained models (3 emojis $\times$ 8 seeds $\times$ 2 update rates).
 
For each trained model, developmental trajectories were generated by running the trained models with random number generator seed used during their training. 
All subsequent analyses are performed on these recorded trajectories, which correspond to the ontogeny of the trained model at inference time.
 
\begin{table}[htbp]
    \centering
    \caption{Mapping of target emojis for GNCA training, including Unicode code points, official CLDR names, and LaTeX shortcodes.}
    \label{tab:emoji_mapping}
    \begin{tabular}{llll}
        \toprule
        Unicode & Name (CLDR) & Shortcode & Emoji \\
        \midrule
        \texttt{U+1F578} & Spider Web & \texttt{spider-web} & \emojispiderweb \\
        \texttt{U+1F603} & \begin{tabular}{@{}l@{}}Grinning Face\\with Big Eyes\end{tabular} & \texttt{smiley} & \emojismiley \\
        \texttt{U+1F98E} & Lizard & \texttt{lizard} & \emojilizard \\
        \bottomrule
    \end{tabular}
\end{table}

\section{Development of Morphology and Structure}
 
Toward an understanding of GNCA development, we begin with the most fundamental question: how does a GNCA reach its target morphology, and how is its internal structure assembled during this process?
One might naively expect that a GNCA approaches its target form gradually and monotonically from the seed state.
However, many dynamical systems that construct complex structures pass through transient states that deviate from such a gradual, direct trajectory.
Biological development, for example, passes through transient states that do not resemble their final forms, with echinoderms as an extreme example.
A GNCA might therefore also exhibit non-monotonic trajectories.

Here the development was examined from two complementary perspectives.
First, the MSE loss between the model output and the target RGBA image was measured over time.
Since the models were trained with this loss, it characterizes the trajectory of morphological convergence to the training target.
Second, the temporal evolution of channel-wise cosine distance matrices was obtained.
These matrices reveal how the hidden representational structure becomes organized in parallel with the visible morphological changes.
 
\paragraph{Morphological convergence.}

\begin{figure}[t]
    \centering
    \includegraphics[width=\columnwidth,angle=0]{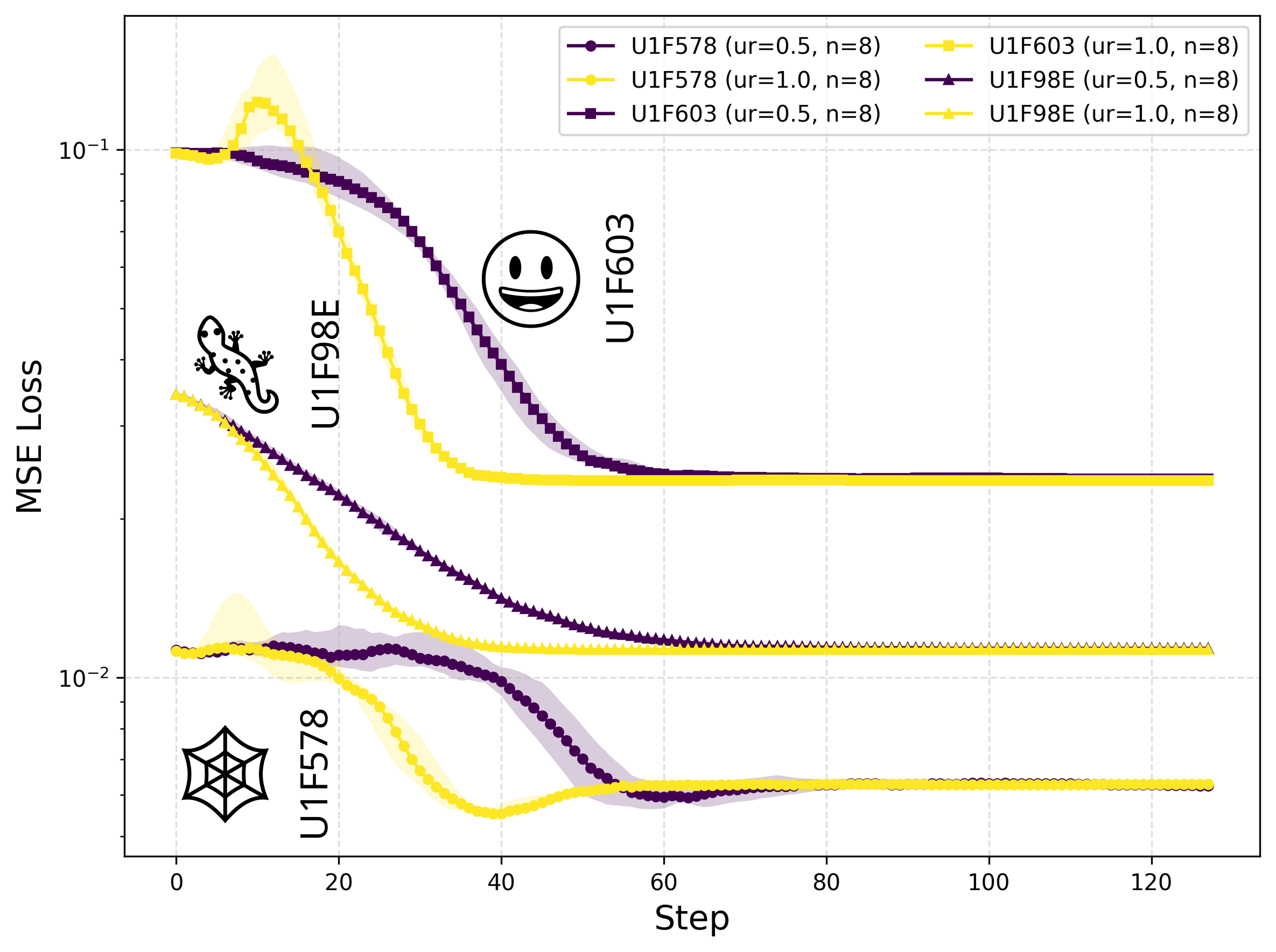}
    \caption{
        MSE loss between developing GNCA and target RGBA image over 128 developmental steps. 
        For all targets ($n=8$), the median and the 25th--75th percentile range are shown.
    }
    \label{fig:onto_loss}
\end{figure}

Figure~\ref{fig:onto_loss} shows the MSE loss between the developing GNCA and the target RGBA image over 128 steps, across all emoji targets and both update rate conditions.
All trained models reached a low final MSE and settled after development, consistent with the loss achieved at the end of training.
This confirmed that the seed cell successfully develops into the target morphology.

The trajectories revealed that morphological convergence is often non-monotonic.
Under synchronous updating (UR~$=1.0$), \texttt{smiley} exhibited a clear overshoot, with the MSE rising above its initial value around Steps 10--15 before declining sharply.
\texttt{spider-web} under the same condition instead showed an undershoot, dropping to a minimum around Steps 35--40 before rising slightly to its final value.
Under asynchronous updating (UR~$=0.5$), \texttt{spider-web} passed through a rugged, temporary plateau until approximately Step 30 before settling to its final value.
The remaining models, including \texttt{lizard} at UR~$=0.5$ and $1.0$ and \texttt{smiley} at UR~$=0.5$, showed a monotonic decrease until they reached a plateau.

Because the MSE was computed over the entire RGBA image, these trajectories characterize the convergence of the morphology as a whole. 
The overshoot and undershoot therefore indicate that the whole form does not always approach the target by incremental refinement. 
In several models, the morphology instead passed through transient intermediate forms that deviated from the target before settling. 
At the level of the whole morphology, GNCA development is therefore non-incremental and proceeds through transient developmental forms rather than monotonic refinement toward the target.

\paragraph{Internal structure.}
To examine the development of the channels of a GNCA from the seed cell, in which all hidden channel values are one, channel-wise cosine distance matrices were computed at exponentially spaced time steps (Steps 1, 2, 4, 8, 16, 32, 64, 128).
For each step, the spatial grid of each channel across the living cells was flattened into a vector and L2-normalized.
Then the pairwise cosine distances were computed between all 16 channels.
Cosine distance was chosen because channels differ substantially in magnitude, and this metric captures differences in spatial pattern independently of scale.
The resulting matrices were reordered by hierarchical clustering (average linkage) for ease of capturing channel-wise differentiation.

\begin{figure}[t]
    \centering
    \includegraphics[width=0.7\columnwidth,angle=0]{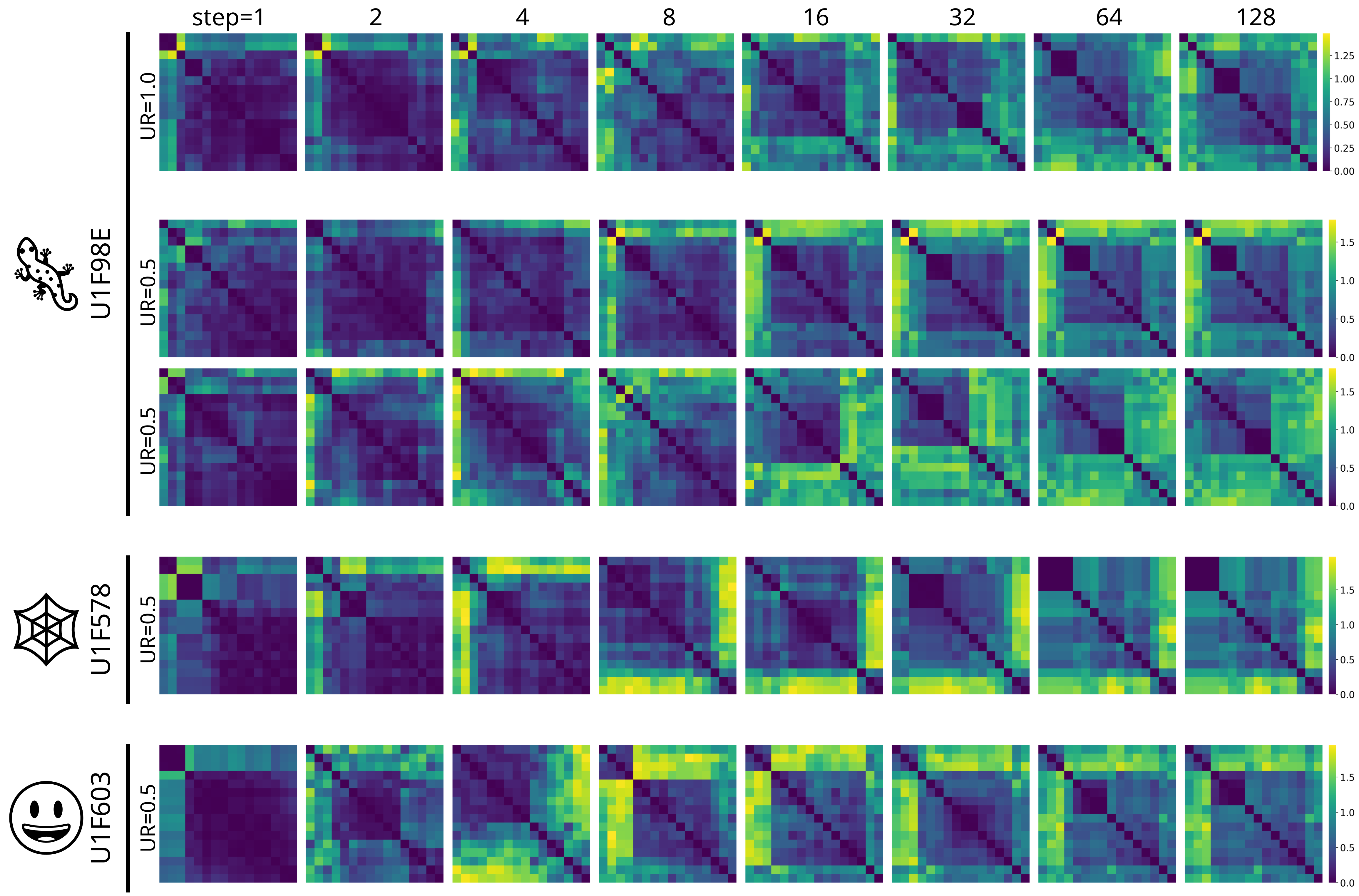}
    \caption{
        Inter-channel cosine distance matrices at developmental steps $[1, 2, 4, 8, 16, 32, 64, 128]$, reordered by hierarchical clustering. \texttt{lizard} is shown for both UR~$= 1.0$ (one random seed) and UR~$= 0.5$ (two seeds, as an example of variability in channel assignment across seeds); \texttt{spider-web} and \texttt{smiley} each show one seed at UR~$= 0.5$.
    }
    \label{fig:ch_cos_dmat}
\end{figure}

The computed matrices are visualized in Figure~\ref{fig:ch_cos_dmat}.
At early developmental steps (Steps 1--4), the distance matrices were largely uniform, reflecting that the channels were still homogeneous.
By Steps 8--16, distinct block structure began to emerge.
This indicates that subsets of channels develop correlated spatial patterns while becoming orthogonal to other channel groups.
This structure stabilized around Steps 32--64 and remained largely unchanged through Step 128.
This suggests that the internal representational architecture is established during the same period in which morphological convergence occurs.

Comparing the two perspectives, the emergence of channel block structure (Steps 8--32) overlapped temporally with the period of most rapid MSE decrease.
This implies that the organization of internal representations and morphological convergence proceed in parallel rather than sequentially.
The specific block structure --- which channels cluster together and how many groups form --- varied across models, even among those trained on the same target with the same update rate but different random seeds (see the two UR~$=0.5$ rows for \texttt{lizard} in Figure~\ref{fig:ch_cos_dmat}).
 
\section{State Space Diversification and Geometry}
 
The preceding section described how GNCA morphology and internal channel structure emerge over time. 
In this section, we investigate how this diversification proceeds. 
First, the expansion of state space through the developmental process was quantified by tracking the total variance and mean pairwise cosine distance among individual living cells as functions of time. 
Second, the intrinsic dimensionality of the state space at each time step was estimated to measure how many effective dimensions the cell states occupy as development progresses. 
Finally, the smoothness of state transitions between time steps was assessed to determine whether the state space possesses a continuous manifold structure. 
This last point serves as a necessary foundation for the subsequent section, where we extract discrete cell types from the continuous state space.

\paragraph{Cell state diversification.}
The diversification of cell states was quantified as the total variance and the mean pairwise cosine distance among cells. 
Let $\mathcal{C}_t = \{i \mid \alpha_i^{(t)} > 0.1\}$ be the set of living cells at each time step $t$, where $\alpha_i^{(t)}$ is the alpha channel value of cell $i$ at step $t$. 
The total variance was defined as $\mathrm{Var}(t) = \sum_{c=1}^{16} \mathrm{Var}(\{s_{i,c}^{(t)}\}_{i \in \mathcal{C}_t})$, which captures the overall spread of cell states across all channels. 
The mean pairwise cosine distance among all living cells was defined as $\bar{d}_{\cos}(t) = \frac{1}{|\mathcal{C}_t|(|\mathcal{C}_t|-1)} \sum_{i \neq j} (1 - \cos(\mathbf{s}_i^{(t)}, \mathbf{s}_j^{(t)}))$, which measures directional diversity independently of magnitude.
 
\begin{figure}[t]
    \centering
    \includegraphics[width=\columnwidth,angle=0]{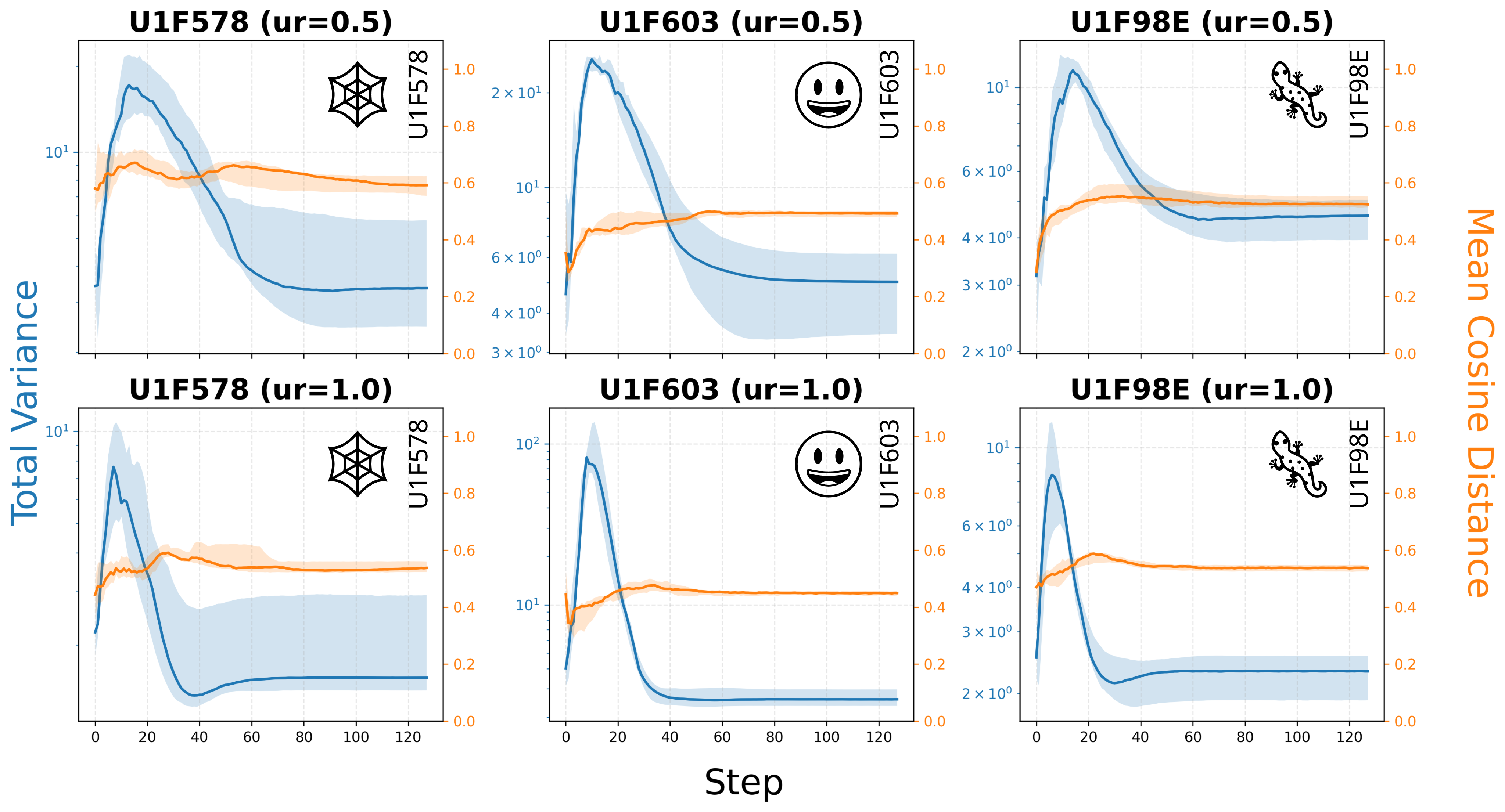}
    \caption{
        Total variance (blue) and mean pairwise cosine distance (orange) among living cells over 128 developmental steps, for UR~$= 0.5$ (top) and UR~$= 1.0$ (bottom). 
        For all targets ($n=8$), median and 25th--75th percentile bands are shown.
    }
    \label{diversity_plot}
\end{figure}

Figure~\ref{diversity_plot} shows both measures across all emoji targets and update rate conditions.
For all targets, where $n=8$ independently trained models are available, the median trajectory and 25th--75th percentile band are shown. 
Several consistent patterns can be identified from this result.
Total variance (blue) exhibited a pronounced peak during early development (approximately Steps 5--15) before declining to a stable plateau. 
This peak occurred slightly earlier under synchronous updating (UR~$= 1.0$) than under asynchronous updating (UR~$= 0.5$). 
This overshoot indicates that cells initially spread widely through state space before contracting to a more compact but structured configuration corresponding to the mature morphology. 
Mean cosine distance (orange) rose rapidly during the first 10--20 steps and then stabilized, which indicates that the directional diversity of cell states is established early and maintained thereafter.
 
The transient peak in total variance was temporally aligned with the non-monotonic MSE trajectories observed in the preceding section.
The period of maximum state space expansion coincided with the period during which the morphology was furthest from its target. 
This temporal coincidence, observed consistently across all models, suggests that the MSE overshoot and the variance peak are manifestations of the same underlying process, a point we return to in the Discussion.
 
A notable difference between update rate conditions was visible. 
Under UR~$= 1.0$, the variance peak was sharp and the subsequent decline steep, whereas under UR~$= 0.5$ the peak was broader and the decline more gradual. 
Across all three targets, the total variance at the stable plateau remained higher under UR~$= 0.5$ than under UR~$= 1.0$. 
This likely reflects that asynchronous updating maintains greater diversity among cell states in the mature morphology.
 
\paragraph{Geometry of the state space.}
The preceding measures quantified how much cell states diversify, but did not reveal the geometric structure of the space they occupy. 
We characterize this structure in terms of dimensionality and smoothness.

To estimate the effective dimensionality available to cell states, the intrinsic dimensionality (ID) was computed at each time step using the maximum likelihood estimator of \citet{Levina2004-dv}. 
For each step $t$, the state vectors $\mathbf{s}_i^{(t)} \in \mathbb{R}^{16}$ of all living cells were extracted and L2-normalized onto the unit hypersphere. 
With these vectors, the ID was estimated using $k = 15$ nearest neighbours. 

\begin{figure*}[t]
    \centering
    \includegraphics[width=0.9\textwidth,angle=0]{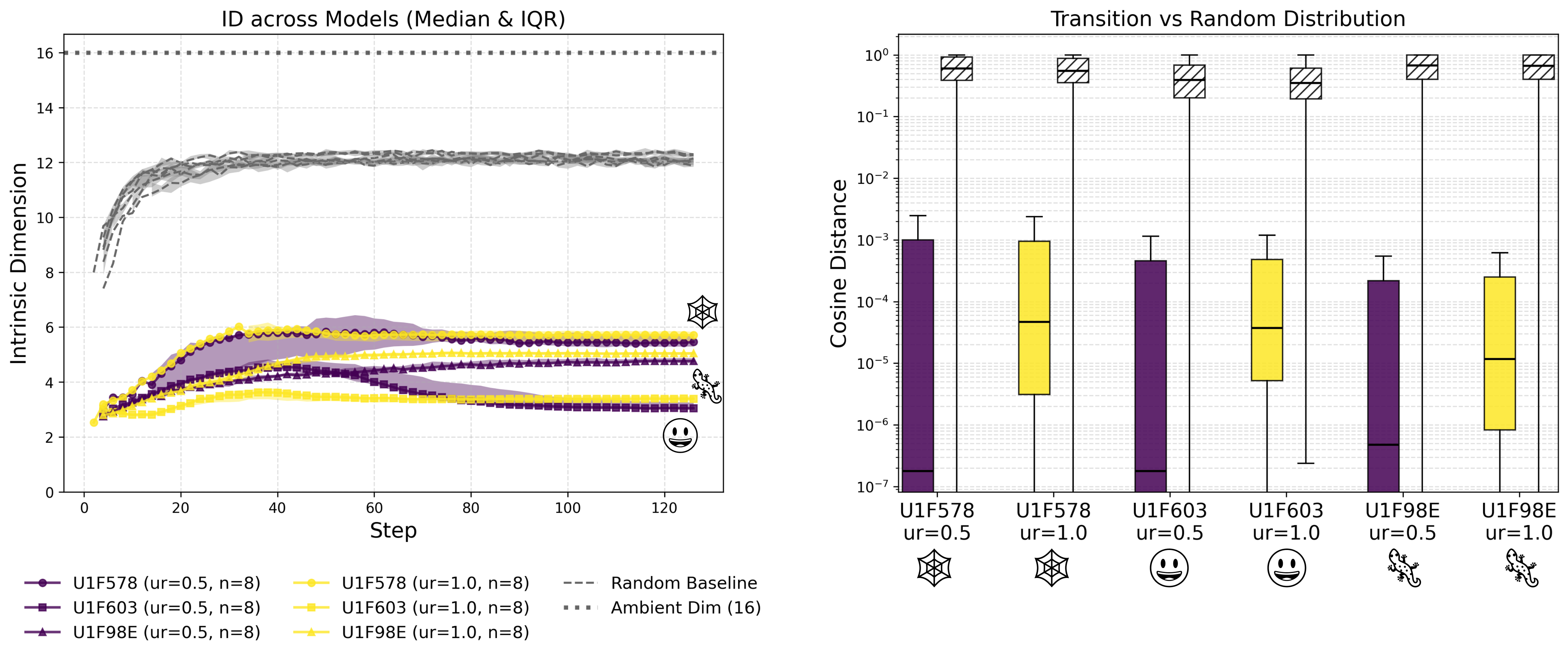}
    \caption{
        Left: intrinsic dimensionality over developmental steps (grey: random baselines; dashed: ambient dimensionality). Right: 9-neighbour minimum cosine transition distances for GNCA models (coloured) versus spatially shuffled baselines (hatched). 
        For all targets ($n=8$), median and 25th--75th percentile are shown.
    }
    \label{manifold_continuity}
\end{figure*}

The left panel of Figure~\ref{manifold_continuity} shows ID trajectories across all models. 
A random baseline of the same size consistently estimated an ID near 12 (grey dash lines).
This confirms that the estimator is well-calibrated.

In contrast, all GNCA models exhibit substantially lower intrinsic dimensionality. 
It began at approximately 2--3 at Step 1 and gradually increased and stabilized at approximately 3--6 by Step approximately 100. 
This confirms that GNCA cell states, despite residing in a 16-dimensional ambient space, are confined to a low-dimensional manifold whose dimensionality grows as development proceeds and cell states diversify. 
After stabilization, the UR = 1.0 models reached slightly higher final ID values than the UR = 0.5 models across all three targets. 
The cause of this difference remains unclear.
Also, among the targets, \texttt{smiley} settled at the lowest final ID.

Low intrinsic dimensionality suggests that cell states lie on a manifold, but does not guarantee that developmental transitions along this manifold are smooth. 
To assess continuity, whether each cell's state at time $t$ can be found in the spatial neighbourhood of a similar state at time $t+1$ was measured. 
Since GNCA cells do not move but information propagates through local interactions \citepalias{Masumori_undated-iu}, a cell's state may be inherited not only by itself but also by its spatial neighbours. 
The present analysis computed this continuity as the minimum cosine distance between its L2-normalized state vector and the L2-normalized state vectors of all 9 cells in the $3 \times 3$ spatial neighbourhood (including itself) at time $t+1$ for each living cell at position $(y, x)$ at time $t$:
 
\begin{equation}
    d_{\mathrm{trans}}(y,x,t) = \min_{(y',x') \in \mathcal{N}(y,x)} \left(1 - \cos\!\left(\hat{\mathbf{s}}_{y,x}^{(t)},\; \hat{\mathbf{s}}_{y',x'}^{(t+1)}\right)\right)
\end{equation}
 
where $\hat{\mathbf{s}}$ denotes L2-normalized state vectors and $\mathcal{N}(y,x)$ is the $3 \times 3$ neighbourhood. 
As a baseline, cell states at time $t+1$ was spatially shuffled to destroy spatial structure while preserving the marginal distribution, and the same 9-neighbour minimum distance was computed. 

The right panel of Figure~\ref{manifold_continuity} compares the distributions of $d_{\mathrm{trans}}$ for GNCA models (coloured boxes) against the spatially shuffled baseline (hatched boxes). 
Across all models and conditions, the transition distances were orders of magnitude smaller than the random baseline (note the logarithmic scale), with median values typically below $10^{-4}$ compared to baseline medians near $10^{-1}$. 
This suggests that state transitions during GNCA development are highly continuous.

Taken together, these results establish that GNCA cell states diversify within a low-dimensional, smooth manifold. 
This structure provides the foundation for the following section, which investigates whether discrete cell types can be identified within this continuous manifold.

\section{Cell Differentiation}

In biological development, cell differentiation refers to the emergence of discrete cell types within this continuum. 
Do analogous discrete cell types emerge in GNCA? 
To address this question, an $\epsilon$-neighbour network ($\epsilon$-NN) was constructed over the cell state vectors, and community extraction was applied to identify communities, which are interpreted as cell types. 
Then the temporal dynamics of these cell types were traced as the number of cells belonging to each community over the course of development, and their spatial distribution on the cell grid was visualized.

\paragraph{Community detection method.}
For each trained model, the L2-normalized 16-dimensional state vectors of all living cells (alpha $> 0.1$) across all developmental time steps were collected into a single set. 
Then, an $\epsilon$-NN graph in which two cells are connected by an edge if and only if the cosine distance between their state vectors is at most $\epsilon$ was constructed.
The edge weights of the graph were defined as the inverse of the distance. 
This graph aggregates cells from all time steps into a unified state space, so that a community may contain cells from different spatial locations and different developmental stages. 
Community structure is extracted using the Louvain algorithm \citep{Blondel2008-vz}, which partitions the graph into communities by maximizing modularity. 
Each community is interpreted as a cell type, that is, a class of cells that share similar internal states and potentially span multiple time steps and spatial positions.
 
\begin{figure}[t]
    \centering
    \includegraphics[width=\columnwidth,angle=0]{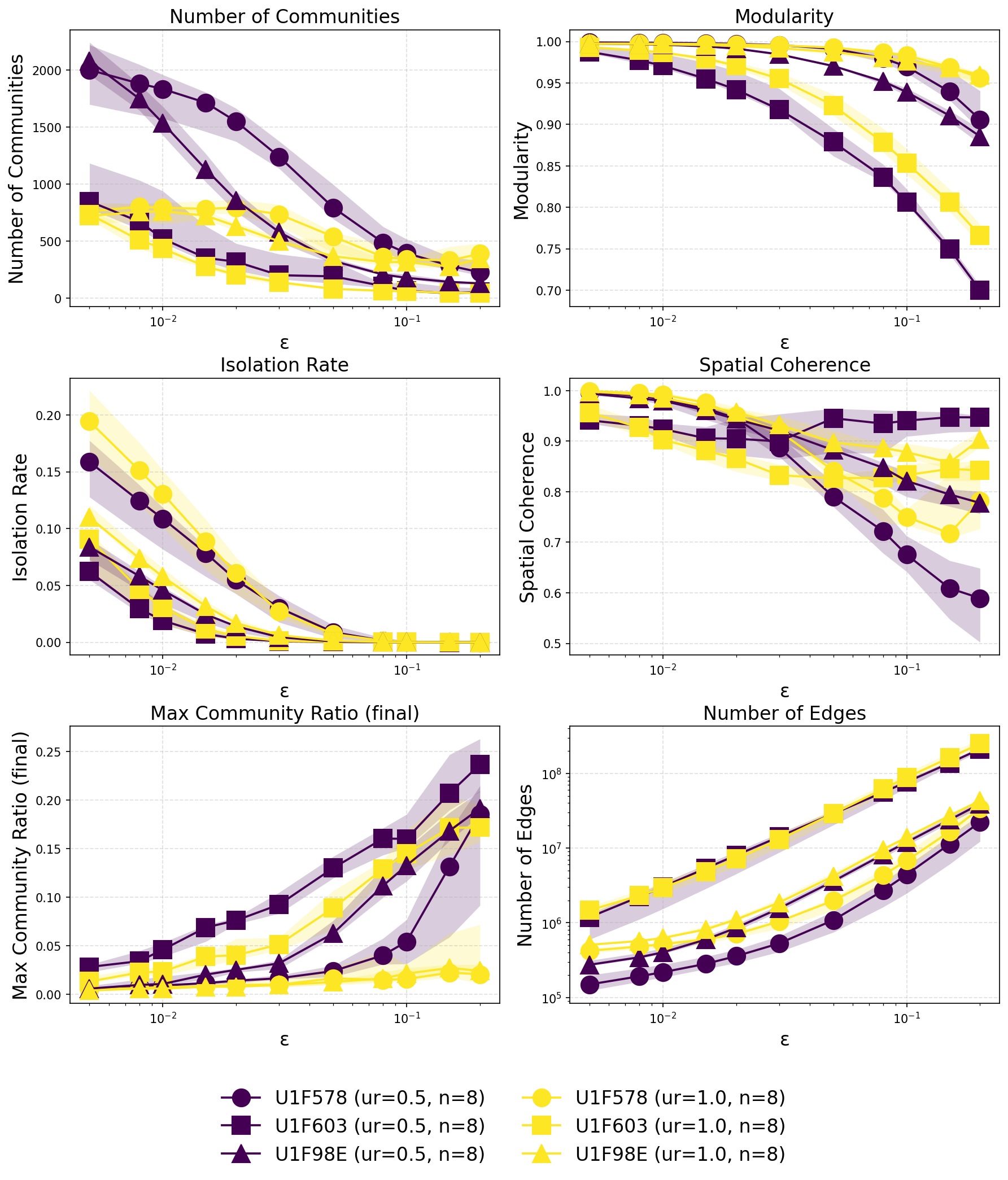}
    \caption{
        Dependence of $\epsilon$-NN community detection metrics on the distance threshold $\epsilon$ across all conditions (three emoji targets $\times$ two update rate conditions $\times$ $n=8$ seeds each). 
        Top row: number of detected communities, Louvain modularity. 
        Middle row: isolation rate, spatial coherence of communities.  
        Bottom row: maximum community fraction at the final developmental step, and total number of graph edges.
    }
    \label{epsilon_search}
\end{figure}

\paragraph{Characterization of the $\epsilon$-NN structure.}
A methodological consideration arises from the fact that the intrinsic dimensionality of the state space increases during development. 
This means that a fixed $\epsilon$ threshold may resolve communities at different effective granularities at early versus late developmental stages. 
In other words, small $\epsilon$ yields many fine-grained communities, while large $\epsilon$ merges cells into fewer, coarser groups. 
To characterize the dependence of the community detection on this parameter and to select an appropriate operating range, $\epsilon$ was swept across the range $[0.005, 0.2]$ for all conditions. 
Figure~\ref{epsilon_search} summarizes the results.
 
Several features were consistent across all conditions. 
The number of detected communities decreased monotonically with $\epsilon$, from over 1,500 at $\epsilon = 0.005$ to fewer than 50 at $\epsilon = 0.2$. 
The Louvain modularity remained above 0.9 for $\epsilon \leq 0.05$ and decreased gradually thereafter. 
The isolation rate (the fraction of cells not assigned to any community) dropped below 5\% by $\epsilon \approx 0.01$ for most models. 
This indicates that the $\epsilon$-NN graph becomes well-connected at this threshold. 
Spatial coherence was measured as the mean fraction of each community's cells belonging to their largest spatially connected component. 
This metric remained above 0.85 for $\epsilon \leq 0.05$ and decreased gradually at larger $\epsilon$ as spatially distant cells merged into the same community.
 
Notable differences emerged between target morphologies. 
\texttt{spider-web} exhibited substantially higher isolation rates at small $\epsilon$ than the other targets, requiring $\epsilon \geq 0.02$ before the graph became well-connected. 
This reflects the narrow distribution of cell states in the spider web morphology, which is nearly monochromatic and spatially repetitive. 
A larger $\epsilon$ is therefore needed to bridge the gaps between these states. 
In contrast, \texttt{smiley} and \texttt{lizard}, which contain more diverse colour regions (eyes, mouth, skin), produced cell states that are more widely dispersed. 
This allows community structure to emerge at smaller $\epsilon$.
 
Taken together, these metrics indicate that $\epsilon = 0.1$ provides a suitable operating point for visualizing community structure. 
At this value the isolation rate dropped to near zero for all models, so few cells remained unassigned. 
At the same time, the modularity remained around 0.9 for all conditions except \texttt{smiley} at UR~$= 1.0$, so the community partition is still well-defined. 
The value $\epsilon = 0.1$ therefore balances these two requirements, and it is used as the default for the visualizations that follow. 
For \texttt{smiley} at UR~$= 1.0$, where the modularity is lower at this value, it was additionally visualized at $\epsilon = 0.005$ to resolve finer cell type structure.
 
\begin{figure}[t]
    \centering
    \includegraphics[width=0.8\columnwidth,angle=0]{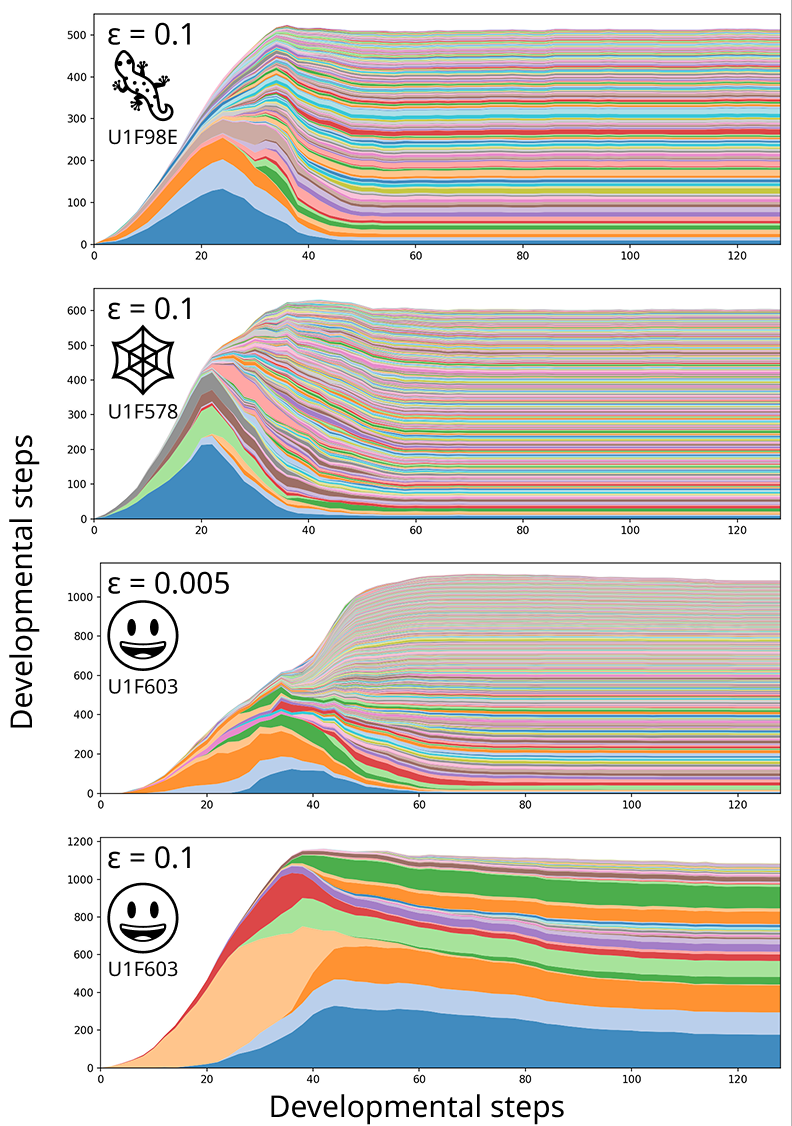}
    \caption{
        Cell type dynamics during GNCA development. Stacked area plots showing the number of cells in each community over 128 developmental steps. 
        Communities detected at $\epsilon = 0.1$ are shown for \texttt{lizard} and \texttt{spider-web} and at $0.005$ and $0.1$ to illustrate that the resolution parameter reveals different scales of cell type organization. 
        Models visualized here are trained under update rate 1.0. 
    }
    \label{cell_differentiation_stack}
\end{figure}

\begin{figure}[t]
    \centering
    \includegraphics[width=\columnwidth,angle=0]{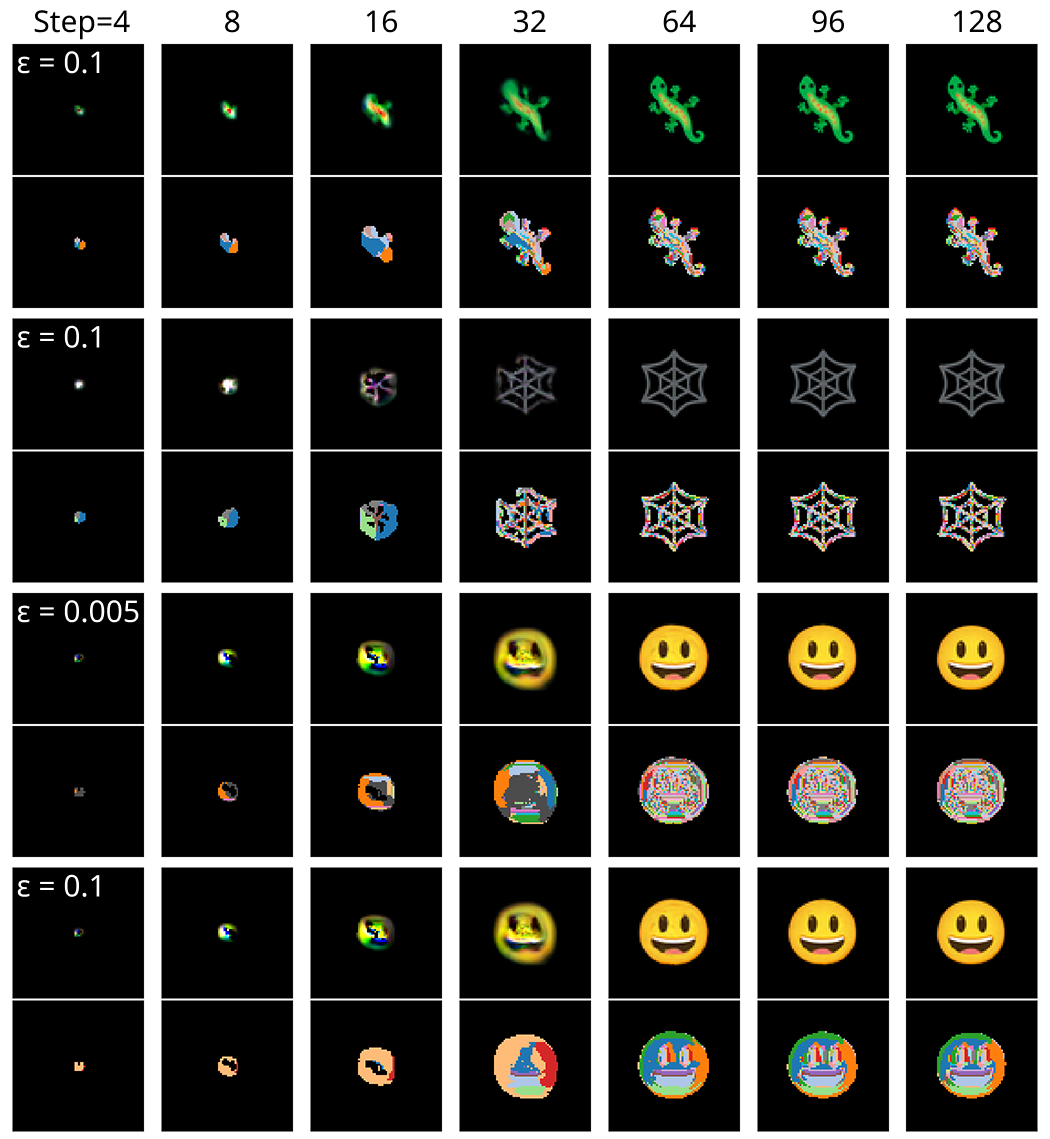}
    \caption{
        Cell type dynamics during GNCA development. RGBA visualization of channels 1--4 (top) and cell grids coloured by their community (bottom) at selected time steps. 
        Communities detected at $\epsilon = 0.1$ are shown for \texttt{lizard} and \texttt{spider-web} and at $0.005$ and $0.1$ to illustrate that the resolution parameter reveals different scales of cell type organization. 
        Models visualized here are trained under update rate 1.0. 
    }
    \label{cell_differentiation_map}
\end{figure}

\paragraph{Developmental dynamics of cell types.}
\cref{cell_differentiation_stack,cell_differentiation_map} show the temporal dynamics of cell communities for all three emoji targets, presented as stacked area plots alongside spatial community maps at selected developmental steps.
 
For \texttt{lizard}, development began with a small number of communities (Steps 0--10). 
Between Steps 10 and 30, several of these initial communities grew rapidly and reached peak cell counts before declining. 
By Step 40--60, the initial large communities had largely been replaced by a more diverse set of smaller, persistent communities. 
The spatial maps showed that these communities first organized into a small number of regions that corresponded to anatomically coherent parts of the lizard morphology. 
Around Step 32, distinct communities mapped to the body, limbs, head, and tail. 
At later steps the number of communities increased further, and these regions were subdivided into smaller communities. 
Once this subdivision was complete, the spatial community structure remained stable for the rest of the observed time range. 
The correspondence between the early large regions and the later fine communities cannot be determined from the present analysis, and this is left to future work.
 
\texttt{spider-web} showed a qualitatively similar developmental trajectory, though the final community structure was less diverse. 
This is consistent with the simple, nearly monochromatic colour of the target image. 
 
For \texttt{smiley}, results at two values of $\epsilon$ are shown to illustrate how the resolution parameter reveals structure at different scales. 
At $\epsilon = 0.1$, coarse-grained communities delineated the major anatomical regions of the smiley face: the yellow skin, the dark eyes, the white mouth region, and the surrounding background each corresponded to a distinct community. 
At $\epsilon = 0.005$, the same regions were resolved into many smaller communities, and the mature morphology appeared mosaic-like. 
These fine communities were not transient. 
They emerged during development and remained stable through the rest of the observed time range.
This indicates that the coarse regions contain finer cell type structure that is itself stable. 

Across both resolutions, and across all models, the fundamental pattern is conserved.
That is, initial communities that dominate early development undergo differentiation into a more diverse set of communities that form the spatial structure of the mature morphology.

\section{Discussion}
 
This study set out to ask what kind of time-developing system a Growing Neural Cellular Automata is. 
By tracing the full developmental trajectory from seed to mature morphology, it has shown that the development of GNCA involves (1) transient intermediate states in the whole morphology, (2) reorganization of channel-wise structures, and (3) the progressive emergence and differentiation of cell types in a continuous state space.
These identified characteristics of GNCA development indicate that the developmental process of GNCA is a reorganization of transient states and structure rather than incremental refinement.

Importantly, these growth dynamics discovered across all three analyses temporally coordinate. 
Specifically, the emergence of channel block structure, the transient variance peak, and the expansion and contraction of transient cell communities all unfold within the same early developmental stage (approximately Steps 5--30).
This coordination implies how GNCA generally develops by providing complementary views on a single process. 
That is, the development of GNCA begins with differentiation into a few strongly distinct cell types, which then subdivide into finer types. 
This is suggested by the findings that the number of early, transient cell types was kept small throughout the early stage while the variance among cell states increased. 
Also, the contraction of these transient cell types began along with the decrease of the variance after the overshoot and the beginning of the channel block formations. 
In the language of state space, the growth of GNCA can be seen as a process in which the cells first differentiate into a few cell types widely separated in the space, and then settle into a larger set of finer-grained attractors in the mature morphology.

One might ask whether these transient phenomena are simply artifacts of the optimization procedure. 
To be sure, training-time choices such as the Adam optimizer and the pool replacement strategy shape the loss landscape and can introduce transient perturbations. 
However, if the MSE overshoot were merely optimizer noise, one would not expect it to coincide with systematic changes in channel block structure, state space dimensionality, and community composition. 
This coordination suggests that the transient phase reflects genuine structural reorganization rather than optimization noise. 
 
These developmental dynamics bear a resemblance to patterns observed in biological ontogeny. 
For instance, the division of the developing body into germ layers and segments might correspond to differentiation of GNCA cells into early cell types. 
However, the parallels should not be overstated because no valid comparison has yet been made.
What is notable is that such a minimal system, with shared local rules, no global coordinator, and no explicit developmental program, spontaneously exhibits staged differentiation and spatial patterning. 
This suggests that these features may be generic properties of developmental systems that construct complex patterns from local interactions. 


A notable feature of GNCA is that the same local rule governs both the developmental process and the dynamics that follow it. 
Once the target spatiotemporal pattern is developed, the same rule continues to operate and maintains it. 
The post-developmental regime of this rule has been shown to exhibit structured internal fluctuations that are functionally necessary for self-repair \citepalias{Masumori_undated-iu}. 
The present analysis has characterized the developmental phase, but how the developmental dynamics connect to this post-developmental regime remains open. 
Clarifying this connection is a natural direction for future work.
 
Several limitations should be noted. 
This analysis is restricted to a single architecture and three simple target morphologies. 
Thus, generalization to more complex settings remains open. 
The observed differences between synchronous and asynchronous updating are suggestive but would benefit from larger sample sizes to be conclusive. 
The $\epsilon$-NN community detection introduces a resolution parameter. 
Robustness was demonstrated across a wide range of $\epsilon$, but density-based methods such as HDBSCAN could address the interaction between fixed $\epsilon$ and increasing intrinsic dimensionality during development. 
Future work could examine the functional role of transient communities through perturbation experiments, investigate how training shapes the developmental trajectory, and extend this framework to GNCA that perform tasks beyond image formation.

\section{Conclusion}
 
We have presented a developmental study of Growing Neural Cellular Automata, revealing that GNCA development is a staged, coordinated process involving non-monotonic convergence, modular channel self-organization, state diversification within a smooth low-dimensional manifold, and the emergence of transient and stable discrete cell types.
These findings position GNCA as a framework for studying how complex developmental processes arise from simple local interactions.
 
\section{Acknowledgements}
This work was supported by JSPS KAKENHI grant No.\ 24H00707, 26K14991, and 26H01260.

The authors used Anthropic Claude and Google Gemini for coding and writing. 
The authors reviewed and edited the content as needed and take full responsibility for the content of the publication.

\footnotesize
\bibliographystyle{apalike}
\bibliography{references}

\end{document}